\documentclass[]{article}

\usepackage{arxiv}

\usepackage{amsmath}
\usepackage{graphicx}
\usepackage{xcolor}
\usepackage{amsfonts}
\usepackage[ruled,vlined,algo2e]{algorithm2e}
\usepackage{amsmath}
\usepackage[colorlinks=true, citecolor=red, linkcolor=red,urlcolor=red,unicode]{hyperref}
\usepackage{lipsum}  
\usepackage{algpseudocode,algorithm}
\usepackage{epstopdf}
\usepackage[british]{babel}
\usepackage{lipsum}

\usepackage[natbibapa]{apacite}
\usepackage{hyperref}
\usepackage{xr}

\newcommand\numberthis{\addtocounter{equation}{1}\tag{\theequation}}

\title{Successor Representation Active Inference}

\begin{document}

\author{
 Beren Millidge \thanks{Corresponding author.} \\
  MRC Brain Networks Dynamics Unit, \\
  University of Oxford, UK, \\
  Verses Research Lab, \\
  Los Angeles, California, USA \\
  \texttt{beren@millidge.name}
   \And
 Christopher L Buckley \\
 Sussex AI Group, Department of Informatics, \\ University of Sussex, UK \\
  Verses Research Lab, \\
  Los Angeles, California, USA \\
  \texttt{C.L.Buckley@sussex.ac.uk}
  }

\maketitle 

\begin{abstract}
Recent work has uncovered close links between between classical reinforcement learning algorithms, Bayesian filtering, and Active Inference which lets us understand value functions in terms of Bayesian posteriors. An alternative, but less explored, model-free RL algorithm is the successor representation, which expresses the value function in terms of a successor matrix of expected future state occupancies. In this paper, we derive the probabilistic interpretation of the successor representation in terms of Bayesian filtering and thus design a novel active inference agent architecture utilizing successor representations instead of model-based planning. We demonstrate that active inference successor representations have significant advantages over current active inference agents in terms of planning horizon and computational cost. Moreover, we demonstrate how the successor representation agent can generalize to changing reward functions such as variants of the expected free energy.

\end{abstract}
\section{Introduction}

Active Inference (AIF) is an unifying theory of action selection in theoretical neuroscience \cite{friston2015active,friston2012active,friston2009reinforcement}. It suggests that action selection, like perception, is fundamentally a problem of inference and that agents select actions by maximizing evidence under a biased generative model \cite{friston2017active,da2020active}. Active inference operates under the aegis of the Bayesian brain hypothesis \cite{knill2004bayesian,doya2007bayesian} and free energy principles \cite{friston2006free,friston2012free,buckley2017free,aguilera2021particular,friston2019physics,millidge2021mathematical} and possesses several neurobiological process theories \cite{friston2017active,parr2019neuronal}. 

Recent work \cite{millidge2020relationship} has uncovered close links between active inference and the framework of control as inference, which shows how many classical reinforcement learning algorithms can be understood as performing Bayesian inference to infer optimal actions \cite{levine2018reinforcement,attias2003planning,rawlik2013probabilistic,toussaint2009probabilistic}. These works, as well as the related duality between control and inference in linearly solvable MDPs \cite{todorov2009efficient,todorov2008general} has allowed us to understand classical objects in reinforcement learning such as Q-functions and  value functions in terms of Bayesian filtering posteriors. Similarly, close connections between active inference and reinforcement learning methods have also been demonstrated \cite{millidge2019combining,millidge2019deep,tschantz2020reinforcement}. It has been shown that deep active inference agents can be derived that can perform actor-critic algorithms \cite{millidge2019deep} as well as model-based reinforcement learning \cite{tschantz2020reinforcement,tschantz2020control,fountas2020deep}, while the fundamental difference between them has been found to be related to the encoding of value into the generative model \cite{millidge2020relationship,millidge2021applications}. Moreover, it has become obvious that active inference can be treated in a model-free (Bellman-equation) paradigm with simply a distinct reward function (the expected free energy) \cite{millidge2019deep,da2020relationship}. However, while much of this work has focused on understanding value functions and model-based RL, another fundamental object in model-free reinforcement learning is the successor representation, which has received much less attention overall. The successor representation (SR) \cite{dayan1993improving} provides an alternative way to estimate value functions. Instead of estimating the value function with the Bellman backup, a successor matrix of long-term discounted state transitions is estimated instead and then dynamically combined with the reward function to yield the value function for a fixed policy. Compared to estimating the value function directly, the SR requires more memory to store the successor matrix but grants the ability to dynamically recompute the value function as the reward function changes as well as providing a compressed form of a `cognitive map' of the environment which can be directly used for exploration and option discovery \cite{machado2017eigenoption,machado2020count,momennejad2020learning}. moreover, from a neuroscientific perspective, the SR has been closely linked to representations in the hippocampus \cite{momennejad2017successor,stachenfeld2017hippocampus} which are concerned with representing abstract (usually spatial) relations \cite{whittington2020tolman,behrens2018cognitive,whittington2022build}.

In this work, applying the probabilistic interpretation of the SR, we showcase how the SR can be directly integrated with standard methods in active inference, resulting in the successor-representation active inference (SR-AIF) agent. We show how SR-AIF has significant computational complexity benefits over standard AIF and that, moreover, the explicit generative model in AIF enables the SR to be computed instantly without requring substantial experience in the environment. Additionally, we show how SR methods can flexibly represent the value function of the EFE and can be used to dynamically trade-off exploration and exploitation at run-time.

\section{Active Inference}

Discrete-state-space active inference possesses a large literature and several thorough tutorials \cite{da2020active,friston2017process,smith2022step}, so we only provide the essentials here. AIF considers agents acting in POMDPs with observations $o$, states $x$, and actions $u$. The agent optimizes over policies $\pi = [u_1, u_2, \dots ]$ which are simply sequences of actions. The agent is typically assumed to be equipped with a generative model of the environment $p(o_{1,T},x_{1:T})$ which describes how observations and states are related over time. This generative model can be factorized into two core components: a likelihood model $p(o_t | x_t)$ which states how observations are generated from states and is represented by a likelihood matrix denoted $A$, and a transition model $p(x_t | x_{t-1}, u_{t-1})$ which states how states change depending on the previous state and action and is represented by a transition matrix denoted $B(u)$. The rewards or goals of the agents are encoded as strong priors in the generative model and are encoded in a `goal vector' denoted $C$.
Since AIF considers agents embedded in a POMDP it has to solve both state inference and action selection problems. State inference is performed using variational inference with a categorical variational distribution $q(x_t)$ which is obtained by minimizing the variational free energy for a specific timestep $t$,
\begin{align}
    q^*(x_t) = \underset{q}{argmin} \, \, \mathcal{F}_t = \underset{q}{argmin} \, \, \mathbb{E}_{q(x_t | o_t)}[\log q(x_t | o_t) - \log p(o_t, x_t | x_{t-1}, u_{t-1})]
\end{align}
AIF uses a unique objective function called the \emph{Expected Free Energy} (EFE) which combines utility or reward maximization with an information gain term which promotes exploration. AIF agents naturally perform both reward-seeking and information-seeking behaviour \cite{friston2015active,parr2019generalised,millidge2021whence,millidge2021understanding}. The EFE is defined as,
\begin{align*}
    \mathcal{G}_t(o_t, x_t) &= \mathbb{E}_{q(o_t, x_t)}[\log q(x_t) - \log \tilde{p}(o_t, x_t |x_{t-1}, u_{t-1})] \\ &=\underbrace{\mathbb{E}_{q(o_t,x_t)}[\log \tilde{p}(o_t)]}_{\text{Expected Utility}} - \underbrace{\mathbb{E}_{q(o_t)}\big[KL[q(x_t | o_t)||q(x_t)]\big]}_{\text{Expected Information Gain}} \numberthis
    \label{EFE_derivation}
\end{align*}
Where $\tilde{p}$ is a `biased' generative model which contains the goal prior vector $C$. As can be seen, the EFE can be decomposed into a reward-seeking and exploratory component which underlies the flexible uncertainty-reducing behaviour of AIF agents. To select actions, AIF samples from the prior over policies $q(\pi)$ which is defined as the softmax over the path integral of the EFE into the future for each timestep. Typically, future policies are evaluated up to a time horizon $T$. This path integral can be expressed as,
\begin{align}
    q(\pi) = \sigma(\sum_t^T \mathcal{G}^\pi_t(o_t, x_t))
    \label{policy_prior}
\end{align}
where $\sigma(x) = \frac{e^{-x}}{\sum_x e^{-x}}$ is the softmax function. Evaluating this path integral exactly for each policy is typically computationally extremely expensive and has exponential complexity due to the exponentially branching number of possible futures to be evaluated. This causes AIF agents to run slowly in practice and has encouraged research into alternative `deep' active inference agents which estimate this path integral in other more efficient (but only approximate) ways \cite{friston2021sophisticated} Here, we present a novel approach based on successor representations.

\section{Successor Representation}

The Successor Representation \cite{dayan1993improving} provides an alternative way to compute the value function of a state. Instead of directly learning the value function (or Q function), for instance by temporal difference (TD) learning, the successor representation learns the \emph{successor matrix}, which is the discounted long term sum of expected state occupancies, from which the value function can be dynamically computed by simply multipling the successor matrix with the reward function. This allows the SR to instantly adapt behaviour to changing reward functions online without explicit model-based planning.

The value function can be defined as the expected long term sum of rewards,
\begin{align*}
    \mathcal{V}^\pi(x) &= r(x) + \gamma B^\pi \mathcal{V}^\pi(x) \\ &= r(x) + \gamma B^\pi [r(x) + \gamma^2 B^\pi [r(x) + \gamma^3 B^\pi[ \cdots]]] \numberthis
    \label{Bellman_equation}
\end{align*}
Where we assume a fixed policy $\pi$ for the value function and transition matrix $B$ and where $\gamma$ is a scalar discount rate. Due to this fixed policy assumption, the max operator in the Bellman equation disappears, so the Bellman equation becomes linear. We can rearrange,
\begin{align*}
    \mathcal{V}^\pi(x) & = r(x) + \gamma B^\pi [r(x) + \gamma^2 B^\pi [r(x) + \gamma^3 B^\pi[ \cdots]]] \\& = (I + \gamma B^\pi + \gamma^2 B^\pi B^\pi + \cdots)r(x) = M^\pi r(x) \numberthis
    \label{SR_derivation}
\end{align*}
Where $M^\pi$ is the successor matrix and can be thought of as encoding the long-run probability that state $x$ transitions to state $x'$.

\section{Successor Representation as Inference}

In a special class of MDPs known as linearly solvable MDPs there is a general duality between control and inference \cite{todorov2008general,todorov2009efficient} such that control can be cast as a Bayesian filtering problem where the value function corresponds to the posterior. To see this, consider the optimal Bellman equation,
\begin{align}
    \mathcal{V}^*(x_t) = \underset{u}{argmax}\big[ r(x_t) + c(u) + \mathbb{E}_{p(x_{t+1} | x_t,u)}[\gamma \mathcal{V}^*(x_{t+1})] \big]
    \label{linear_rl}
\end{align}
where we have added an additional control cost $c(u)$. The fundamental challenge is the nonlinearity of the Bellman equation due to the argmax operation. \cite{todorov2009efficient} noticed that if the dynamics are considered to be completely controllable and set by the action $p(x_{t+1} | x_t, u) = u(\cdot | x_t)$, while the original dynamics instead take the form of a `dynamics prior' in the control cost which is set to $KL[u(\cdot | x_t)||p(x_{t+1} | x_t, u)]$ which penalizes divergence from the prior dynamics, then the argmax is analytically solvable.  By defining the `desirability function' $z(x) = e^{-\mathcal{V}^*(x)}$ and exponentiating, we can obtain a linear equation in $z$,
\begin{align}
    z(x) = e^{-r(x)}\mathbb{E}_{p(x_{t+1}| x_t)}[\gamma z(x_{t+1})]
    \label{desirability_posterior}
\end{align}
which can be solved easily. Crucially, however, this equation takes the same form as the Bayesian filtering recursion $p(x_t | o_t) \propto p(o_t | x_t)\mathbb{E}_{p(x_t | x_{t-1}}p(x_{t-1} | o_{t-1})$ when we make the identification of the `desirability' $z(x_t)$ with the posterior $p(x_t | o_t)$ and the exponentiated reward $e^{-r(x_t)}$ with the likelihood $p(o_t | x_t)$. Interestingly, this same relationship between exponentiated reward and probability is also used heuristically in the control as inference literature \cite{levine2018reinforcement}. An additional subtle point is that control is about the future instead of the past so the sum telescopes forward instead of backwards in time. By factoring Equation \ref{linear_rl} as in Equation \ref{SR_derivation}, it is straightforward to observe that,
\begin{align}
    M = \sum_{\tau=t}^T \gamma^\tau \Big( \prod_{i=t}^T \sum_{x_i} p(x_i | x_{i-1})\Big) = \sum_{\tau = t}^T \gamma^\tau p(x_\tau | x_t)
    \label{M_probabilistic_interpretation}
\end{align}
In effect, we can think of $M$ as representing the discounted sum of the probabilities of all the possible times to reach state $x$ over the time horizon. A similar and novel result can be derived for the case of general (not linearly solvable) MDPs but with a fixed policy except here we derive an upper bound on the SR instead of an equality. We begin by taking the log of the backwards Bayesian filtering posterior and then repeatedly applying Jensen's inequality to obtain,
\begin{align*}
    \log p(x_t | o_t) &= \log p(o_t | x_t) + \log \mathbb{E}_{p(x_{t+1} | x_t)}[\log p(x_{t+1} | o_{t+1})] \\
    &\leq \log p(o_t | x_t) + \mathbb{E}_{p(x_{t+1} | x_t)} \Big[ \big[ \log p(o_t | x_t) + \mathbb{E}_{p(x_{t+2} | x_{t+1})}[ \log \dots ] \big] \Big] \numberthis
    \label{jensens_derivation}
\end{align*}
Which has the same recursive structure as the linear Bellman Equation for a fixed policy (Equation \ref{Bellman_equation}) so long as we maintain the equivalence between the value function and the log posterior and the reward and the log likelihood and implicitly set the discount factor $\gamma$ to 1. The technique in Equation \ref{SR_derivation} can then be applied to give the same probabilistic interpretation of the SR as Equation \ref{M_probabilistic_interpretation}. In sum, we have shown how optimal control can be associated with filtering and Bayesian posteriors exactly in the case of linear MDPs and the Bayesian posterior as an upper bound in the case of a fixed policy. These results provide a sound probabilistic and Bayesian interpretation of the SR, which has hiterto been missing in the literature, and let us design a principled active inference agent based upon the SR. 
\section{Successor Representation Active Inference}
Using the probabilistic interpretation of the SR and the equations of discrete state-space AIF, we can construct an AIF agent which utilizes the SR to compute value functions of actions instead of model-based planning. That is, the policy posterior path integral $q(\pi) = \sigma(\mathcal{G}) = \sigma(\sum_t^T \mathcal{G}_t)$ can be considered as a value function and dynamically computed using the SR. The fact that in discrete AIF the generative model transition matrix $B(u)$ is given allows us to dispense with learning the successor matrix from experience. However, to apply the SR, we need to choose which policy $\pi$ we wish to compute the value function under. This choice is important since the default policy must assign enough probability mass to all parts of the state-space to be able to provide an accurate value estimate there. Heuristically, we set the default policy to be uniform over the action space $p(u) = \frac{1}{\mathcal{A}}$ where $\mathcal{A}$ is the cardinality of the action space. This lets us define the default transition matrix,
\begin{align}
    \tilde{B} = \mathbb{E}_{p(u)}[B(u)] = \frac{1}{\mathcal{A}}\sum_i B[:,:,u_i]
    \label{average_B}
\end{align}
Given $\tilde{B}$, we can analytically calculate the SR using the infinite series result,
\begin{align}
    M^\pi = (I + \gamma \tilde{B} + \gamma^2 \tilde{B}^2 \cdots) = (I - \gamma \tilde{B})^{-1}
    \label{analytical_SR}
\end{align}
This means that as long as the generative model is known, the EFE value function $q(\pi)$ can be computed exactly without any interaction with the environment by first computing $M^\pi$ as in Equation \ref{analytical_SR} and then multiplying by the reward function which is the EFE $\mathcal{G} = M^\pi \mathcal{G}^\pi(x)$. From this EFE value function actions can be sampled from the posterior over actions as,
\begin{align}
    u \sim q(\pi) = \sigma(\mathcal{G})
    \label{action_sampling}
\end{align}
A slight complication is that while the SR is defined for MDPs, AIF typically assumes a POMDP structure with observations $o$ that do not fully specify the hidden state but are related through the likelihood matrix $A$. We address this by computing observation value functions as the expected state posterior under the state posterior distribution,
\begin{align}
    \mathcal{V}^\pi(o) = \mathbb{E}_{q(x | o)}[\mathcal{V}^\pi(x)] = q \mathcal{M}^\pi \mathcal{G}^\pi
    \label{observation_VF}
\end{align}
where $q = [q_1, q_2 \cdots]$ is the categorical variational posterior. The SR-AIF algorithm can thus be summarized as follows: we are given a generative model containing the $A$ and $B$ matrices and a set of desired states $C$. At initialization, the agent computes the successor matrix $M^\pi$ using the default policy with Equation \ref{average_B}. For each action in a given state, SR-AIF computes the EFE value function $\mathcal{G}^\pi$ for that action and then actions are sampled from the policy posterior which is the softmax over the EFE action-value functions. In a POMDP environment exactly the same process takes place except instead of action-state we have action-observation value functions which are computed as Equation \ref{observation_VF}.
\subsection{Computational Complexity}
In theory the computational complexity of SR-AIF is superior than standard AIF as standard-AIF uses model-based planning which evaluates the EFE value function $\mathcal{G}$ by exhaustively computing all possible future trajectories for different policies. This has a cost that grows exponentially in the time horizon due to the branching of possible futures. If we denote the number of actions $\mathcal{A}$, the dimension of the state-space $\mathcal{X}$ and the time-horizon $T$, we can approximately say that the computational complexity of standard AIF is of order $\mathcal{O}(\mathcal{X}T^2 \cdot \mathcal{A}^T)$ since the number of possible trajectories is approximately $\mathcal{A}^T$ where evaluating each step of a trajectory costs of order $\mathcal{X}$ and we must repeat this for each timestep. This is exponential in the time-horizon and renders AIF unsuitable for long term planning. Several heuristic methods have been proposed to handle this, usually by pruning obviously unsuccessful policies \cite{friston2021sophisticated}. However, this does not remove the exponential complexity but only reduces it by a constant factor. In practice, this exponential explosion is also handled by simply reducing the time-horizon or the policy space to be searched, which renders the evaluation of $\mathcal{G}$ approximate and makes the AIF agents myopic to long term reward contingencies.

By contrast, SR-AIF analytically computes an approximation to the EFE value function $\mathcal{G}$ directly from the known transition dynamics by Equation \ref{analytical_SR}. This means that no exhaustive future simulation is required for each action but instead only a one-time cost is incurred at initialization. The main cost is the matrix inverse of approximately $\mathcal{X}^3$. Then an action must be selected which costs of order $\mathcal{A}$. This means the total complexity of SR-AIF is of order $\mathcal{O}(\mathcal{X}^3 + \mathcal{A}T)$. SR-AIF thus reduces the computational complexity of AIF from exponential to cubic and hence, in theory, allows discrete-state-space active inference to be applied to substantially larger problems than previously possible.
\begin{figure}
    \vspace{-0.7cm}
    \centering
    \includegraphics[width=\linewidth]{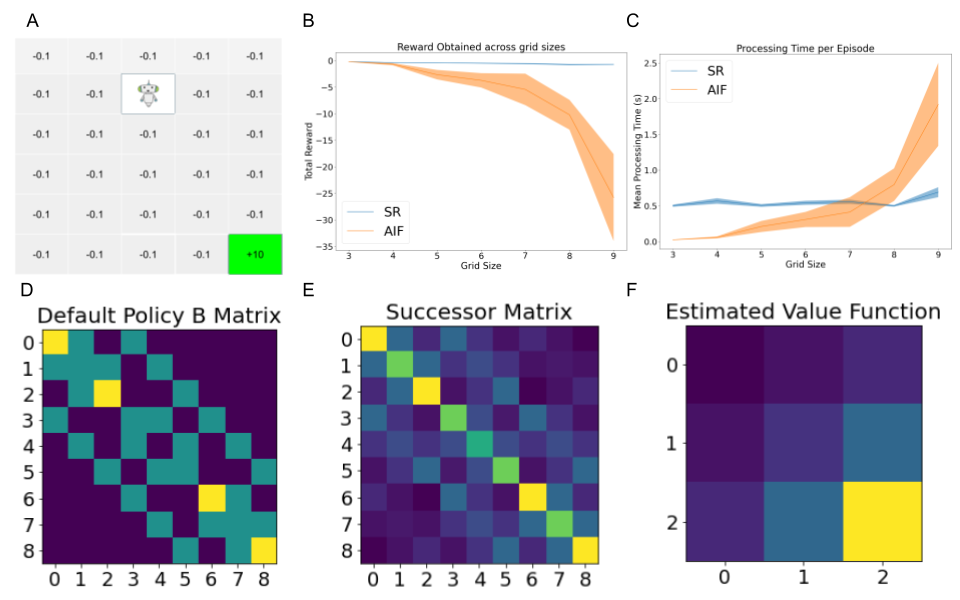}
    \vspace{-0.3cm}
    \caption{Top Row: A: Schematic of the grid-world task. The AIF agent is initialized in a random square and must make it to the bottom corner to obtain reward. B: The total reward obtained on average for SR-AIF and AIF agents. Due to a limited planning horizon, the AIF agent cannot solve larger gridworlds and hence incurs large negative rewards. C: Computational cost (measured in compute time per episode) for SR-AIF. For small gridworlds, SR-AIF is more expensive since the matrix inversion cost dominates while for larger gridworlds the cost of standard AIF increases exponentially. Bottom Row: Visualization of the default policy matrix $\tilde{B}$, Successor matrix $M$, and estimated value function $\mathcal{V}^\pi$ for a $3 \times 3$ gridworld. }
    \label{figure_1}
    \vspace{-0.8cm}
\end{figure}
\section{Experiments}
We empirically demonstrate the superior computational complexity and ultimate performance of SR-AIF as the state-space and time-horizon grows on a series of grid-world environments. These provide a simple test-bed environment for evaluating computational complexity in practice without the confounding factors introduced by a more complex environment. The agent is initialized randomly in an $N \times N$ grid and must reach a reward located in the bottom corner of the grid. On average, as the grid size increases, both the state-space size and the planning horizon required to find this reward increase. 

We implemented the AIF agent using the \texttt{pymdp} library for discrete state-space AIF \cite{heins2022pymdp}. We found that for larger grid-sizes the matrix inverse used to compute the successor matrix often became numerically unstable. Heuristically, we countered this by increasing the `discount rate' $\gamma$ in Equation \ref{analytical_SR} to be greater than $1$ (we used $5$ for larger grid-sizes). Otherwise $\gamma$ for SR-AIF and standard AIF was set to $0.99$. Intuitively this can be seen as weighting future states more than the present and had the numerical effect of increasing the differences in the value function between far away states which would otherwise collapse to a negligible value. This, however, is a heuristic device and removes the probabilistic interpretation of the discount rate as a probability of agent survival as in \cite{levine2018reinforcement}. For the active inference agent, to keep computation times managable, we used an planning horizon of $7$ and policy length of $7$.

This task had no epistemic contingencies but only involved reward maximization. A key aspect of active inference though is its native handling of uncertainty through the EFE objective. Here, we demonstrate that the SR representation can adapt to uncertainty and dynamically change the balance of exploration and exploitation. To demonstrate this, we introduce an uncertainty and exploration component into the gridworld task by setting some squares of the grid to be `unknowable' such that if the agent is on these squares, it is equally likely to receive an observation from any other square in the same row of the grid. This is done by setting columns of the A matrix to uniform distributions for each `unknowable' square in the grid. We show that if equipped with the EFE objective function, SR-AIF is able to instantly recompute the value function based on this information in the A matrix without having to change the successor matrix $M$. Moreover, due to the property that the value function can be recomputed for each reward function, this allows a dynamic weighting of the utility and information gain components of the EFE to take place at runtime.

\begin{figure}
    \centering
    \includegraphics[width=\linewidth]{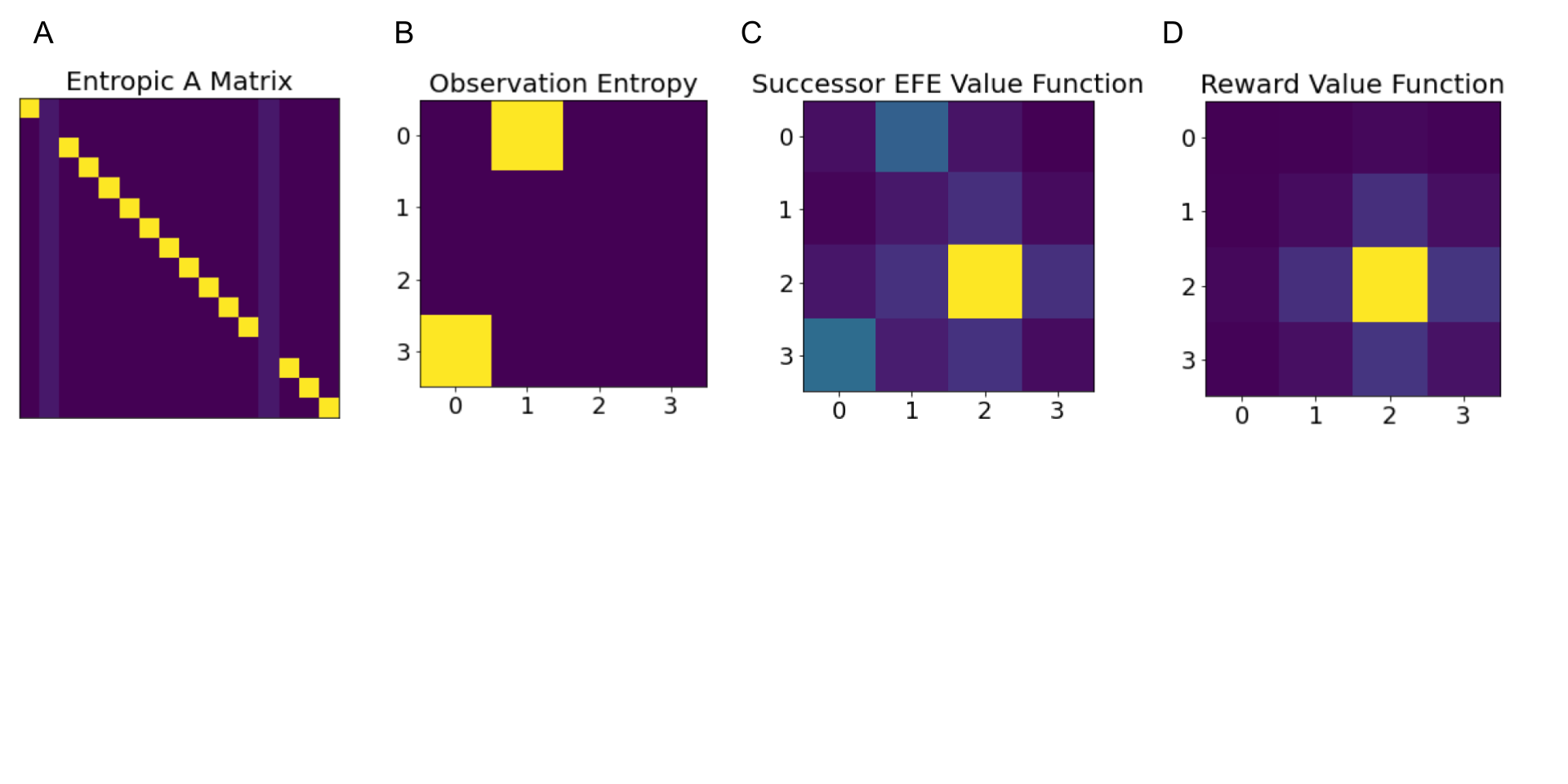}
    \vspace{-3.5cm}
    \caption{Effect of introducing observation uncertainty into the model. A: the A matrix with two `unknowable' squares resulting in a uniform distribution in two columns. B: the corresponding entropy of the state-space with the two `unknowable' squares having high entropy. C and D: The value function computed using the EFE which responds positively to regions of high uncertainty since there is the potential for information gain compared to the standard reward function. SR-AIF is able to correctly combine both utility and epistemic drives at runtime.}
    \label{figure_2}
\end{figure}
\vspace{-1cm}
\section{Discussion}
In this paper, we have derived a probabilistic interpretation of the SR and related it to control as inference and linear RL. We then constructed an SR-AIF algorithm which exhibits superior significant performance and computational complexity benefits to standard AIF due to its amortization of policy selection using a successor matrix which can be computed analytically at initialization. 

It is important to note that while the SR-AIF has substantially better computational complexity, this comes at the cost of a necessary approximation. The successor matrix is computed only for a fixed default policy $\pi$ and the choice of this policy can have significant effects upon the estimated value function and hence upon behaviour. The choice of the default policy is thus important to performance and was here chosen entirely on heuristic grounds. Principled ways of estimating or bootstrapping better default policies would be important for improving the performance of SR-AIF in practice. This is especially important due to the reflexivity of RL environments whereby the default exploration policy determines what data will be sampled from the environment which is then used to further train the model and refine the SR. Alternatively, the MDP itself could be regularized so that it becomes linear as in \cite{todorov2009efficient} such that the optimal policy can be solved for directly. This approach has been applied in a neuroscience context \cite{piray2021linear} but the extension to active inference remains to be investigated.

Another point of extension is that here we have considered AIF and SR-AIF in the context of a single sensory modality and a single-factor generative model. However, many tasks modelled by AIF use multiple factors and modalities to express more complex relationships and contingencies. The extension of SR-AIF to multiple modalities and factors is straightforward algebraically, but has subtle implementation details and is left to future work.

\section{Code Availability}
Code to reproduce all experiments and figures can be found at: \newline
\text{https://github.com/BerenMillidge/Active\_Inference\_Successor\_Representations}.

\section{Acknowledgements}
Beren Millidge is supported by the BBSRC grant BB/S006338/1 and by Verses Research. CLB is supported by BBRSC grant number BB/P022197/1 and by Joint Research with the National Institutes of Natural Sciences (NINS), Japan, program No. 01112005.

\bibliography{cites.bib}

\end{document}